\def\year{2020}\relax
\documentclass[letterpaper]{article} 
\usepackage{aaai20}  
\usepackage{times}  
\usepackage{helvet} 
\usepackage{courier}  
\usepackage[hyphens]{url}  
\usepackage{graphicx} 
\usepackage{graphicx}  
\usepackage{amsmath,amssymb}
\usepackage{multirow}

\title{TextScanner: Reading Characters in Order for Robust Scene Text Recognition}

\author{Zhaoyi Wan\textsuperscript{\rm 1}\footnotemark[\value{footnote}]\thanks{Authors contribute equally},
Minghang He\textsuperscript{\rm 2}\footnotemark[\value{footnote}],
Haoran Chen\textsuperscript{\rm 3},
Xiang Bai\textsuperscript{\rm 2}$^\dagger$,
Cong Yao\textsuperscript{\rm 1}\thanks{Corresponding author}\\
\textsuperscript{\rm 1}Megvii,
\textsuperscript{\rm 2}Huazhong University of Science and Technology,
\textsuperscript{\rm 3}Beijing Institute of Technology\\
i@wanzy.me, \{minghanghe09,xbai\}@hust.edu.cn, xidianchr@163.com, yaocong2010@gmail.com
}

\begin{document}
\maketitle
\begin{abstract}
Driven by deep learning and a large volume of data, scene text recognition has evolved rapidly in recent years. Formerly, RNN-attention-based methods have dominated this field, but suffer from the problem of \textit{attention drift} in certain situations. Lately, semantic segmentation based algorithms have proven effective at recognizing text of different forms (horizontal, oriented and curved). However, these methods may produce spurious characters or miss genuine characters, as they rely heavily on a thresholding procedure operated on segmentation maps. To tackle these challenges, we propose in this paper an alternative approach, called TextScanner, for scene text recognition. TextScanner bears three characteristics: (1) Basically, it belongs to the semantic segmentation family, as it generates pixel-wise, multi-channel segmentation maps for character class, position and order; (2) Meanwhile, akin to RNN-attention-based methods, it also adopts RNN for context modeling; (3) Moreover, it performs paralleled prediction for character position and class, and ensures that characters are transcripted in the correct order. The experiments on standard benchmark datasets demonstrate that TextScanner outperforms the state-of-the-art methods. Moreover, TextScanner shows its superiority in recognizing more difficult text such as Chinese transcripts and aligning with target characters.
\end{abstract}

\section{Introduction}\label{introduction}

In the past decades, scene text detection and recognition have drawn considerable attention from the computer vision community, due to its wide applications, e.g, automatic driving~\cite{ctc}, visual auxiliaries~\cite{visual_attn}, and human-computer interaction~\cite{wang2012end}. As scene text provides pivotal and specific information, accurate recognition of text plays crucial roles in various real-world scenarios~\cite{quy2013recognizing}.

Among the state-of-the-art methods for scene text recognition, there are two prevalent paradigms: RNN-attention-based methods and semantic segmentation based algorithms. The former~\cite{shi2016robust,aon}, drawing inspiration from neural machine translation~\cite{bahdanau2014neural}, encodes images into features and employs an attention mechanism to align and decode characters. The latter~\cite{mask_textspotter,ca-fcn}, approaching text recognition from a 2D perspective, first adopts a fully convolutional network (FCN) to perform semantic segmentation, then seeks connected components in the segmentation maps, and finally infers the class of each connected component (each is taken as a character).

\begin{figure}[t]
\centering
\includegraphics[width=0.97\linewidth]{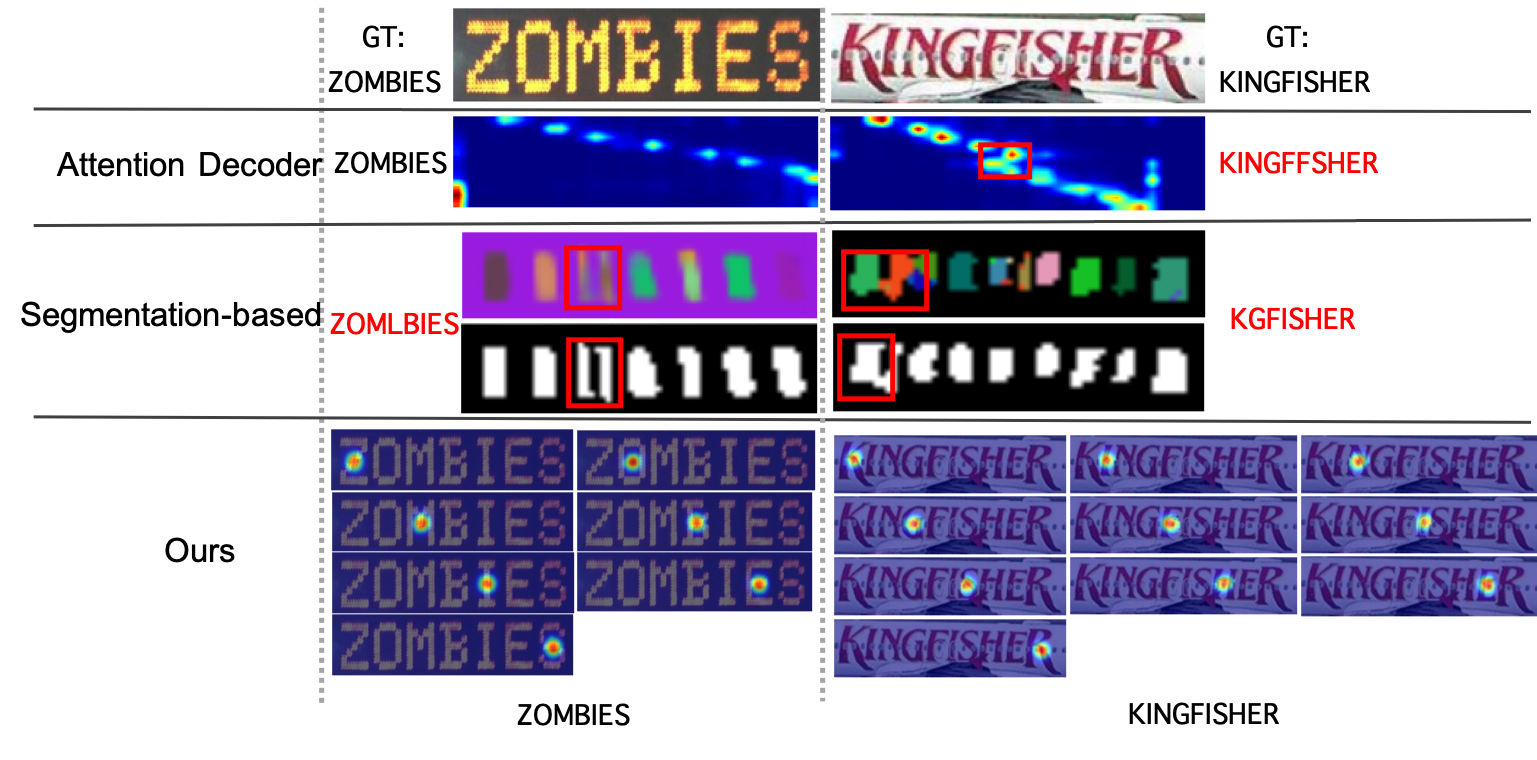}

\caption{Our Motivation.
RNN-attention-based methods may encounter the problem of \textit{attention drift}~\cite{cheng2017fan} (see the red rectangle), thus leading to incorrect prediction of character class.
In semantic segmentation based algorithms, the search of connected components depends on a thresholding operation, which is prone to over-segmentation or under-segmentation, thus generating spurious characters or missing genuine characters (see the red rectangles).
In contrast, TextScanner scans characters one by one and ensures that characters are read in right order and separated properly.}

\label{fig:motivation}
\end{figure}
\begin{figure*}[ht]
\centering
\includegraphics[width=0.8\linewidth]{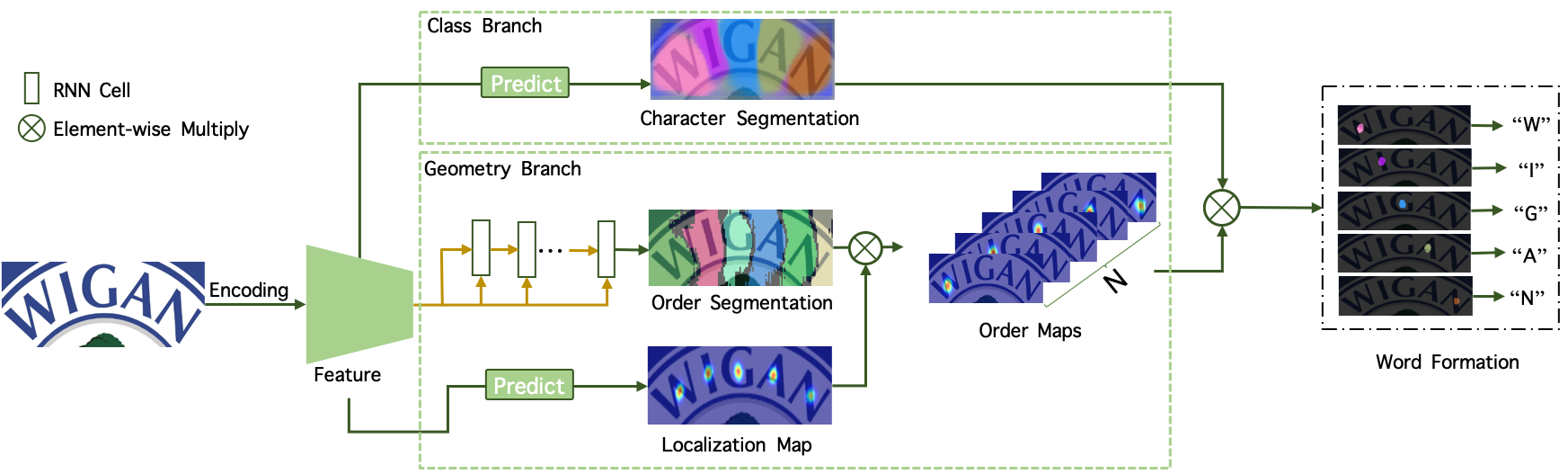}

\caption{Schematic illustration of the proposed text recognition framework. Different colors in character segmentation map represent the values in different channels. The values in the localization map and order maps are visualized as heat maps. The predictions of the two branches are fused to extract characters (position, order, and class) and form the final output.}
\label{fig:framework}
\end{figure*}

\subsubsection{Motivation}

Essentially, to correctly recognize the content in a cropped text image, the number of characters as well as the order and class of each character should be accurately predicted. RNN-attention based methods usually work well in most cases. However, when there is
noise in the background or irregular text shape
, the attention mechanism may fail that the center of the estimated attention map targets to a wrong position, causing mistakes in character order and class (see Fig.~\ref{fig:motivation}). More seriously, due to the recurrent memory mechanism in the RNN module, such errors will accumulate and propagate, making the situation even worse.

Semantic segmentation based algorithms explore a different way and exhibit stronger adaptability to text of different shapes (horizontal, oriented and curved). However, it is difficult to successfully separate each character from the segmentation maps, since improper binarization will result in such embarrassments: one character might be split into multiple parts or multiple characters may stick together (see Fig.~\ref{fig:motivation}). In these cases, the predictions of number and class of characters would be wrong. In summary, existing approaches, either RNN-attention based or semantic segmentation based, are not able to commendably resolve the difficulties in scene text recognition.

The root cause for the attention drift problem in the RNN-attention based methods might be that the alignment operations (realized with attention maps) rely on both visual features and previous decoding results. Mutual interference might occur between these two types of information. Therefore, it is necessary to perform character alignment and classification in independent branches. Regarding semantic segmentation based algorithms, the assumption that characters can be sought via simple binarization does not hold in challenging scenarios. To address this issue, a natural and feasible solution is to represent the position and order of characters with different channels.

\subsubsection{Our Work}

In this paper, we propose a novel text recognition framework, called TextScanner. Like a scanner in the real world, TextScanner can read characters in correct order.

Generally, TextScanner is built upon semantic segmentation~\cite{ca-fcn}. It consists of two branches: one for character classification (class branch) and the other for character position and order prediction (geometry branch) (see Fig.~\ref{fig:framework}). The class branch produces multi-channel segmentation maps, in which the values at each location represent the probabilities of character classes (including the background class). The geometry branch also produces multi-channel segmentation maps, but the meanings of the values at each location are different from those in the class branch. The characters are sought by multiplying two groups of segmentation maps in an element-wise manner and acquiring the class with maximal probability for each channel. This procedure is termed as word formation in this work.

Since characters are well aligned and the order is ensured, TextScanner can avoid the attention drift phenomenon observed in RNN-attention based methods. Meanwhile, in the geometry branch different characters, even contiguous with each other or with the same class label, are strictly assigned into different channels, so they can be easily extracted.

As FAN~\cite{cheng2017fan} and CA-FCN~\cite{ca-fcn}, TextScanner also requires character level annotations for training, since the geometry branch takes character centers as supervision signals. However, there are actually plenty of real image examples without character level annotations, which could be very beneficial for training text recognizers. To make use of such real data, we devise a mutual supervision mechanism. For image examples without character level annotations, the predictions of the class branch and the geometry branch can supervise each other with only sequence level annotations. In consequence, TextScanner is able to fully utilize all kinds of available training data, including both synthetic and real text images.

We conduct experiments on public benchmarks for scene text recognition to validate the proposed TextScanner. It achieves higher or highly competitive accuracy on regular text datasets and obtains significantly enhanced performance on irregular text datasets. The recognition accuracy increases $3.3\%$ on ICDAR 2015 and $4\%$ on CUTE80, compared with the previous art. We also evaluate TextScanner on a Chinese recognition task. The quantitative results further prove the superiority of the proposed algorithm. The contributions in this work are summarized as follows:
\begin{description}
\item[$\bullet$] We propose a novel text recognition framework, which predicts the class and geometry information (position and order) of characters with two separate branches.
\item[$\bullet$] We devise a mutual-supervision mechanism, which endows the framework with the ability to make use of both synthetic and real data for training.
\item[$\bullet$] The experiments demonstrate that the proposed TextScanner achieves state-of-the-art or highly competitive performance on public benchmarks.
\item[$\bullet$] Furthermore, TextScanner exhibits stronger adaptability to longer and more complex text (such as Chinese scripts).
\end{description}

\section{Related Work}

Text recognition has been a long-standing research topic in computer vision. Research efforts on text recognition can date back to the early age of AI~\cite{herbert1982history,lecun1998gradient}. With the rise of deep learning, scene text recognition has entered a new era. For recent progress in this field, please refer to the survey paper~\cite{long2018scene,zhu2016scene}. In this section, we will concentrate on the most relevant works.

Inspired by speech recognition and natural language processing, CTC-based~\cite{shi2017end,ctc} soft-alignment methods and attention-based~\cite{visual_attn,aster} methods are proposed to handle text recognition as a sequence recognition task. Among them, attention-based methods are prevalent recently, and achieve state-of-the-art performance on public benchmarks. On the other hand, there are still challenging problems in the field of scene text recognition. Text images in natural scenes suffer from the complex background, arbitrary text shape and severe image distortion. Most of current text recognition algorithms are not robust enough to solve hard cases such as text instances which are oriented, curved or extremely blurred.

As stated in Sec.~\ref{introduction}, the problem of attention drift is exacerbated by recursive modeling. This problem is also observed in speech recognition~\cite{attention-drift}, which is where the idea of attention decoder originated from.
There are existing works aimed at alleviating this problem. \cite{cheng2017fan} proposed to correct attention positions using characters' class and localization label. New loss function motivated by the formulation of edit distance is presented\cite{bai2018edit} to improve the hard alignment of attention decoder. However, these methods do not change the nature of error accumulating which lies in the coupled modeling of attention generation and character classification. Different from these methods, the proposed algorithm uses two separated branches to classify characters and predict the positions and order of characters. The potential mutual interference between alignment and decoding is eliminated and the problem of attention drift can be avoided.

Recently, segmentation-based methods are also introduced to the field of text recognition~\cite{mask_textspotter,ca-fcn}. Segmentation-based methods are usually more flexible than attention decoders in the recognition of irregular text such as oriented or curved text instances. However, the post-processing of these methods may fail to separate closely arranged characters as shown in Fig.~\ref{fig:motivation}. As the characters are recognized by finding and voting inside the connected components in the segmentation map, this restriction limits their recognition accuracy. Besides, the application of segmentation-based methods remains limited because these methods can not use real image examples with only sequence-level annotation. With our proposed method, the characters are naturally separated and ordered by dispatching character localization to different channels. The mutual-supervision mechanism further enables the two branches to utilize sequence level annotations to supervise and enhance each other. 

\section{Methodology}

\subsection{Overview}
The overall structure of the proposed method is illustrated in Fig.~\ref{fig:framework}. The decoder of the network is composed of two branches: class branch and geometry branch.

\subsubsection{Class Branch}
The class branch of TextScanner produces character segmentation $G \in R^{h \times w \times c}$, which is of resolution $h \times w$ and $c$ denotes the number of classes (all character classes plus background). $G$ is directly generated from visual features extracted by a CNN backbone. The prediction module is composed of two stacked convolutional layers with kernel size $3 \times 3$ and $1 \times 1$. A Softmax normalization is applied over the class dimension to generate the character segmentation maps.

\subsubsection{Geometry Branch} 

\begin{figure}[t]
\centering
\includegraphics[width=0.97\linewidth]{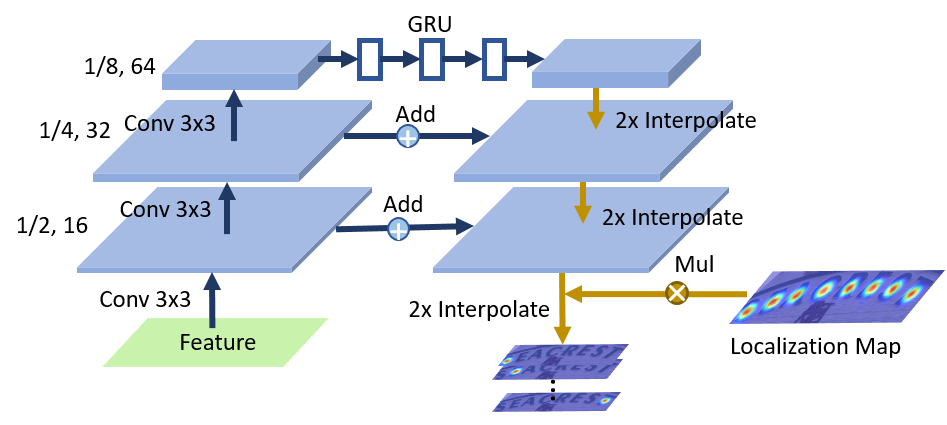}
\caption
{Illustration of the geometry branch. The feature maps are up-sampled and down-sampled by a pyramid architecture with skip connections. Features at the top layer is processed by an RNN module for context modeling.}
\label{fig:order}
\end{figure}

Firstly, a character localization map $Q \in R^{h \times w}$ is produced from the same visual features as the class branch, with a Sigmoid activation function. Concurrently, a top-down pyramid structure, in which features in the upsampling path is added by the features with the same resolution from the downsampling path is taken to generate order segmentation maps. Especially, the feature maps in the top layer of the downsampling path are encoded by an RNN (GRU\cite{GRU}, to be exact) module for context modeling. Following the upsampling path, two convolutional layers are employed to generate the order segmentation maps $S \in R^{h \times w \times N}$, where $N$ is the pre-defined max sequence length. The order segmentation maps are also normalized by a Softmax operation. Then an order map $H_{k} \in R^{h \times w}$, which indicates the position of the $k_{th}$ character in the sequence, can be computed from the $k_{th} (k \in [1, N])$ channel of order segmentation and the character localization map $Q$ by an element-wise multiplication:$H_{k} =Q * S_{k}$.
The detail of the geometry branch is depicted in Fig.~\ref{fig:order}.

\subsubsection{Word Formation}

With the produced character clarification maps $G$ and order maps $H$, we now present the decoding procedure which formats characters in order. As the order of character locations is encoded into the order maps $H$, the classification scores can be computed from the order maps and character clarification maps as:
\begin{equation}
    p_{k} = \int_{(x,y) \in \Omega}{G(x, y) * H_{k}(x,y)}
\end{equation}
where $p_{k} \in R^{c}$ is the vector of scores representing the class probabilities of the $k_{th}$ character. $\Omega$ is all valid spatial locations in the $h \times w$ space. Similar to attention decoders, once the maximal probability of a character is below a pre-defined threshold $T_{score}$ or $k$ reached the maximal value $N$, the decoding process is terminated.

This decoding procedure is totally differentiable. Therefore, it can be trained within the network using sequence level as well as character level annotations. The optimization process utilizing sequence level annotations is introduced in detail in the following section.

\subsection{Pre-Training with Character-Level Annotations}
When pre-training on synthetic data, TextScanner can be optimized with character-level annotations.

\subsubsection{Label Generation}

\begin{figure}[t]
    \centering
    \includegraphics[width=0.97\linewidth]{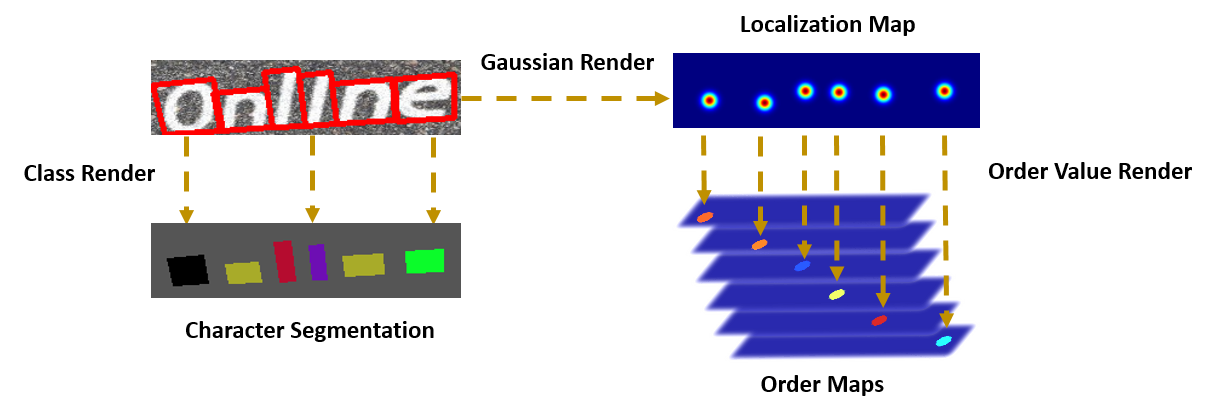}
    \caption{Ground truth generation for pre-training. Pixels outside shrunk boxes $P'$ are represented as gray in character segmentation label, which are ignored in loss computation.}
    \label{fig:labelgeneration}
\end{figure}

Due to the rectangles are inaccurate 
in curved or dense text, we keep the definition of character regions polygons
$P=\{(x_{i}, y_{i})\}_{i=1}^{N_{p}}$, where $N_{p}$ is the number of points in polygon.

To refrain the overlap caused by edges of adjacent characters, the polygon character bounding box $P$ is shrunk to $P'$ with the Vatti clipping algorithm\cite{vati}. Inside $P'$ area, the class of the corresponding character is rendered as ground truth of the character segmentation. Pixels outside $P'$ hardly contribute to the decoding of sequences and are ignored in the optimization of text segmentation.

To generate the ground truth of order maps with character-level annotations, the center of Gaussian maps is firstly detected by computing the central points of characters bounding boxes. As Fig.~\ref{fig:labelgeneration} shown, 

2D Gaussian maps $\hat{Y}_{k} \in R^{h \times w}$ with $\sigma$ and expectation at central points are generated for each character. Then the order of characters is rendered for pixels inside $\hat{Y}_{k}$ area:
\begin{equation}
    \hat{Z}_{k}(i,j) = 
    \begin{cases}
    k, & \textrm{if $\frac{\hat{Y}_{k}(i,j)}{max\hat{Y}_k(h,w)} > \zeta_{order}$, $(i,j) \in (h,w)$} \\
    0, & \text{otherwise}
    \end{cases}
\end{equation}
where $\hat{Z}_{k} \in R^{h \times w}$ is the generated order map ground truth for each character, $\zeta_{order}$ is the order threshold, which set to 0.5 in our experiment. 
Finally $\hat{Z}_{k}$ is normalized to $[0, 1]$, to produce the ground truth $Z_{k}$ of $H_{k}$.

Taking the same operation as $\hat{Z}_{k}$ to ${Z}_{k}$, 0-1 normalized Gaussian heatmap ${Y}_{k}$ can be acquired from $\hat{Y}_{k}$. According to all ${Y}_{k}$, the ground truth of localization map $Q$ can be generated by straightforwardly combining heat maps:
\begin{equation}
    Y^{*} = \max_{k=1}^{N}{Y_{k}}
\end{equation}

\subsubsection{Loss Function}

The overall loss function 
is a weighted sum of losses for the three mentioned tasks as
\begin{equation}
    L = \lambda_{l} * L_{l} + \lambda_{o} * L_{o} + \lambda_{m} * L_{m} + L_{s}
\end{equation}
where $L_{c}$, $L_{o}$, $L_{s}$, $L_{m}$ are the losses for localization map, order segmentation, text segmentation and mutual supervision loss respectively. The detail of mutual supervision loss is illustrated in the next section. In our experiments $\lambda_{l}$ and $\lambda_{o}$ are set to 10 for scaling the numerical values. $\lambda_{m}$ is set to 0 during pretraining otherwise to 1.

The localization map loss is computed as an average smooth l1 loss. The losses for order segmentation($L_{o}$) and character segmentation($L_{s}$) are computed as cross entropy between the predicted scores and corresponding ground truth. The background class in both segmentation task is ignored in cross entropy computation.

\subsection{Mutual-Supervision Mechanism} \label{weakly-supervise}

\begin{figure*}[t]
\centering
\includegraphics[width=0.8\linewidth]{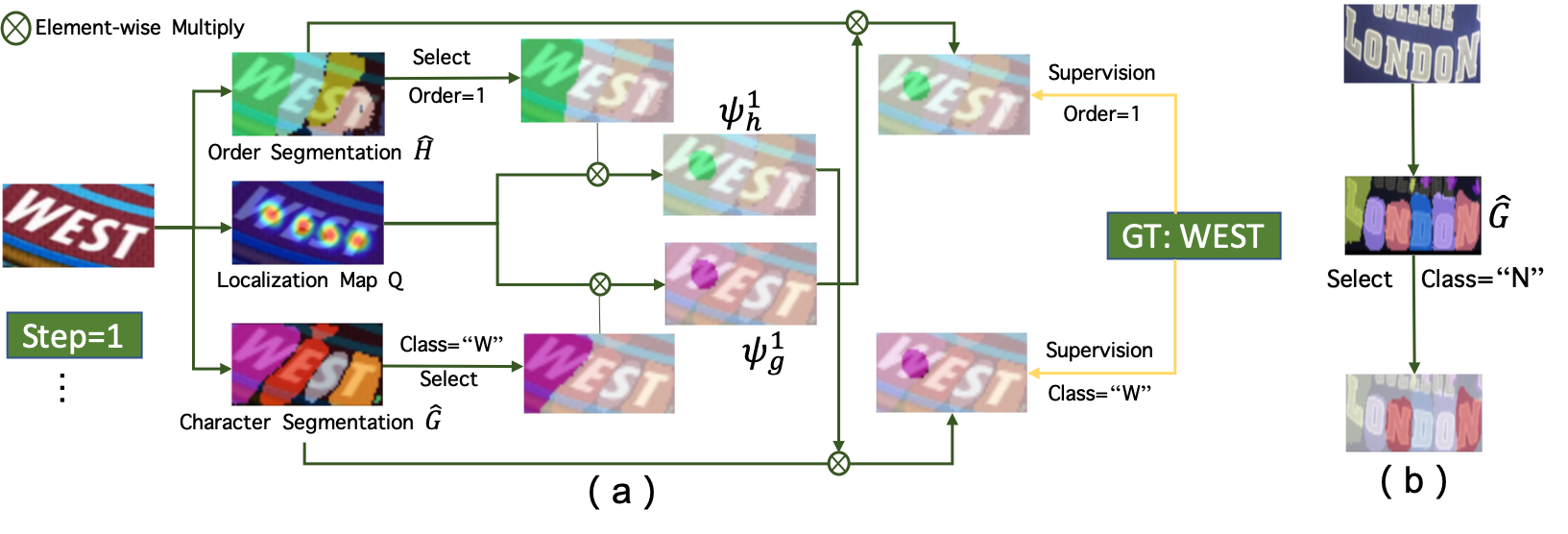}
\caption{(a) Visualization of step 1 of mutual-supervision mechanism. The selected regions in $\hat{G}$ and $\hat{H}$ are refined using $Q$ to get ${\Psi}_{g}^{1}$ and ${\Psi}_{h}^{1}$, which are then mapped into $\hat{H}$ and $\hat{G}$ separately. (b) Two regions in $\hat{G}$ are selected for `N' in ``LONDON".}
\label{fig:mutual supervision}
\end{figure*}

 For previous semantic segmentation based methods, it's critical to obtain accurate locations of all the characters during model training, since the character classification is at pixel-level.  This is problematic when character level annotations are not available.  To reduce the reliance on character-level annotations, we devise a mutual-supervision mechanism based on the dual-branch structure of TextScanner. As shown in Fig.~\ref{fig:framework}, text sequences can be generated by combining character segmentation maps $G$ and order maps $H$. Given a sequence label and one of the two outputs, supervision signals can be generated for the other one.

 ${G}$ and ${H}$ are further transformed by taking the index of max value across $c$ and $o$ channels into $\hat{G} \in R^{h \times w}$ and $\hat{H} \in R^{h \times w}$. Given the text sequence label $T$, the mutual-supervision process is carried out from the first character in $T$ to the last. For the $k^{th}$ character in $T$, its order is $k$ and the class is $T(k)$. ${\Psi}_{h}^{k}$ and ${\Psi}_{g}^{k}$ are coordinates of the pixels that corresponding to the $k^{th}$ character in $H$ and $G$:
\begin{equation}
    \begin{split}
    &{\Psi}_{h}^{k} = \{(i,j) | \hat{H}(i,j) = k, Q(i,j)>\epsilon  \} \\
    &{\Psi}_{g}^{k} = \{(i,j) | \hat{G}(i,j) = T(k),  Q(i,j)>\epsilon\}
    \end{split}
\end{equation}

Apart from the constraint of class or order, we add the constraint of $Q(i,j)>\epsilon$ to keep the attended regions in the center of characters. $\epsilon$ is set to 0.2 in our experiment.

 For the mutual supervision purpose, we use ${\Psi}_{g}^{k}$, which is derived from $G$,in the supervision of $H$ and ${\Psi}_{h}^{k}$, which is derived from $H$, in the supervision of $G$:
  \begin{align}
 \begin{split}
 &L_{g}^{k} = \frac{1}{|{\Psi}_{h}^{k}|}\sum_{(i,j) \in {\Psi}_{h}^{k}}L_{CE}(G(i,j), onehot(T(k))) \\
 &L_{h}^{k} = \frac{1}{|{\Psi}_{g}^{k}|}\sum_{(i,j) \in {\Psi}_{g}^{k}}L_{CE}(H(i,j), onehot(k))
 \end{split}
 \end{align}
 
$L_{CE}$ is the cross entropy loss and $onehot(\cdot)$ is the one hot encoding function. The step 1 of cross supervision process is shown in Fig \ref{fig:mutual supervision}(a). The process is carried on till the last character of $T$. Note that multiple regions of $\hat{G}$ are chosen in one selection for characters occur more than once in $T$ and can't be used in supervision of $H$, as shown in Fig \ref{fig:mutual supervision}(b). So we remove these samples from the cross supervision process. The confidences for $G$ and $H$ are denoted as ${\Phi}_{g}$ and ${\Phi}_{h}$:
\begin{align}
 \begin{split}
&n_{g}^{k}=\begin{cases}1,  & \text{if} \ {\Psi}_{g}^{k} \neq \text{\O} \\\
0, & \text{otherwise}
 \end{cases}, \quad
 \Phi_{g} = \frac{\sum_{k=1}^{|T|}n_{g}^{k}}{|T|}\\
 &n_{h}^{k}=\begin{cases}1,  & \text{if} \ {\Psi}_{h}^{k} \neq \text{\O}  \\
 0, & \text{otherwise}
 \end{cases}, \quad
 \Phi_{h} = \frac{\sum_{k=1}^{|T|}n_{h}^{k}}{|T|}
 \end{split}
 \end{align}
 
 For the $k^{th}$ character in $T$, if the number of pixels in its corresponding region in $G$ or $H$ is larger than 0, the character is considered exists in $G$ or $H$. $|T|$ is the length of $T$. The more the characters in $T$ exists, the higher $\Phi$ is. $T=1$ when all the characters exists in the prediction map. $\Phi_{g}$ and $\Phi_{h}$ are included in the loss computation of the text sequence:

\begin{equation}
\begin{gathered}
 L_{m} = L_{h} + \lambda*L_{g} \\
 L_{g} = \frac{(\Phi_{h})^\gamma}{|T|}\sum_{k=1}^{|T|}L_{g}^{k}, \quad
 L_{h} = \frac{(\Phi_{g})^\gamma}{|T|}\sum_{k=1}^{|T|}L_{h}^{k} 
\end{gathered}
\end{equation}

 In our experiment, $\lambda$ is set to 0.2, $\gamma$ is set to 2 to further reduce the impact of inaccurate predictions. After being pre-trained on synthetic datasets with character level annotations, our model can be further fine-tuned on real-world datasets or synthetic datasets with only sequence level annotation by adopting this cross supervision mechanism, which is impossible in previous segmentation-based methods. 
 
\subsection{Other Details}

Our model is built on top of the backbone from CA-FCN, in which the character attentions are removed and VGG blocks are replaced with a ResNet-50\cite{resnet} base model. The score threshold $\zeta_{score}$ is set as 0.3 empirically, and the max size $N$ is set as 32 in our implementations.

\section{Experiments}

\begin{table*}[t]
    \begin{center}
    \begin{tabular}{|c|c|c|c|c|c|c|c|c|c|c|c|c|}
    \hline
        \multirow{2}{*}{\textbf{Methods}} & \multirow{2}{*}{\textbf{Training Data}} & \multicolumn{3}{|c|}{IIIT} & \multicolumn{2}{|c|}{SVT} & IC13 & IC15 & SVTP & CT \\
        \cline{3-11}
        & & 50 & 1k & 0 & 50 & 0 & 0 & 0 & 0 & 0  \\ \hline
        \hline
        ~Almaz{\'a}n et al.\tiny\cite{almazan2014word} & - & 91.2 & 82.1 & - & 89.2 & - & - & - & - & - \\
        ~Strokelets\tiny\cite{yao2014strokelets} & - & 80.2 & 69.3 & - & 75.9 & - & - & - & - & - \\
        ~Jaderberg et al.\tiny\cite{jaderberg2014deep} & - & - & - & 86.1 & - & - & - & - & - & -\\
        ~Su et al.\tiny\cite{su2014accurate} & - & - & - & - & 83.0 & - & - & - & - & - \\
        ~Gordo\tiny\cite{gordo2015supervised} & - & 93.3 & 86.6 & - & 91.8 & - & - & - & - & - \\
        ~Jaderberg et al.\tiny\cite{jaderberg2016reading} & 90k & 97.1 & 92.7 & - & 95.4 & 80.7 & 90.8 & - & - & - \\
        ~Jaderberg et al.\tiny\cite{jaderberg2014deep} & 90k & 95.5 & 89.6 & - & 93.2 & 71.7 & 81.8 & - & - & - \\
        ~CRNN\tiny\cite{shi2017end} & 90k & 97.6 & 94.4 & 78.2 & 96.4 & 80.8 & 86.7 & - & - & - \\
        ~RARE\tiny\cite{shi2016robust} & 90k & 96.2 & 93.8 & 81.9 & 95.5 & 81.9 & 88.6 & - & 71.8 & 59.2 \\
        ~$R^2$\text{AM\tiny\cite{lee2016recursive}} & 90k & 96.8 & 94.4 & 78.4 & 96.3 & 80.7 & 90.0 & - & - & - \\
        ~Yang et al.\tiny\cite{yang2017learning} & 90k & 97.8 & 96.1 & - & 95.2 & - & - & - & 75.8 & 69.3 \\
        ~CA-FCN\tiny\cite{ca-fcn} & ST & 99.8 & 98.8 & 91.9 & \textbf{98.8} & 86.4 & 91.5 & - & - & 79.9 \\ \hline
        TextScanner-pre & ST & 99.8 & 99.0 & 92.1 & 97.7 & 89.4 & \textbf{92.7} & 79.6 & 82.6 & 83.5 \\
        TextScanner-mutual & ST & \textbf{99.8} & \textbf{99.2} & \textbf{92.6} & 98.4 & \textbf{89.5} & 92.5 & \textbf{80.1} & \textbf{83.7} & \textbf{85.2} \\
        \hline
        \hline
        
        $~FAN^{\ast}$\text{\tiny\cite{cheng2017fan}} & ST + 90k & 99.3 & 97.5 & 87.4 & 97.1 & 85.9 & 93.3 & 70.6 & - & - \\
        $~AON^{\ast}$\text{\tiny\cite{aon}} & ST + 90k & 99.6 & 98.1 & 87.0 & 96.0 & 82.8 & - & 68.2 & 73.0 & 76.8 \\
        $~EP^{\ast}$\text{\tiny\cite{bai2018edit}} & ST + 90k & 99.5 & 97.9 & 88.3 & 96.6 & 87.5 & \textbf{94.4} & 73.9 & - & - \\
        $~ASTER^{\ast }$\text{\tiny\cite{aster}} & ST + 90k & 99.6 & 98.8 & 93.4 & 97.4 & 89.5 & 91.8 & 76.1 & 78.5 & 79.5 \\
        \hline
        TextScanner+90k$^{\ast }$ & ST + 90k & \textbf{99.7} & \textbf{99.1} & \textbf{93.9} & \textbf{98.5} & \textbf{90.1} & 92.9 & \textbf{79.4} & \textbf{84.3} & \textbf{83.3}  \\ 
        \hline
        \hline
        ~SAR$^\dagger$\text{\tiny\cite{li2019show}} & ST + 90k + real & 99.4 & 98.2 & 95.0 & 98.5 & 91.2 & 94.0 & 78.8 & \textbf{86.4} & 89.6 \\
        \hline
        TextScanner+real$^\dagger$ & ST + 90k + real & \textbf{99.8} & \textbf{99.5} & \textbf{95.7} & \textbf{99.4} & \textbf{92.7} & \textbf{94.9} & \textbf{83.5} & 84.8 & \textbf{91.6} \\
    \hline
    \end{tabular}
    \end{center}
    \caption{Performance comparison of our methods and others. ``ST'', ``90k'', and ``real'' are the training data of SynthText, 90k, and real data, respectively.
    The methods marked with star mix SynthText and 90k dataset for training and methods marked with ``$\dagger$" use the training set of real data. ``0", ``50" and ``1k" indicate the size of the lexicons, ``0" means no lexicon.}
    \label{tab:performance}
\end{table*} \label{performance}

We conduct experiments on standard benchmarks to evaluate TextScanner and compare it with other competitors.
\subsection{Datasets} \label{datasets}
\textbf{ICDAR 2013(IC13)}~\cite{ic13} recognition task provides 288 scene images with annotations, from which 1015 word images are cropped. Besides, the dataset provides character-level bounding boxes.

\noindent\textbf{ICDAR 2015(IC15)}~\cite{karatzas2015icdar}
consists of 1000 images with word-level quadrangles annotation for training and 500 for testing.

\noindent\textbf{IIIT 5K-Words(IIIT)}~\cite{mishra2012scene} dataset contains 5K word images for scene text recognition. 

\noindent\textbf{Street View Text(SVT)}~\cite{wang2011end} dataset
has 350 images and only word-level annotations are provided.

\noindent\textbf{SVT-Perspective(SVTP)}~\cite{quy2013recognizing} dataset contains 639 cropped images for testing. Many images in the dataset are heavily distorted. 

\noindent\textbf{CUTE80(CT)}~\cite{cute} dataset is taken in natural scene. It consists of 80 high-resolution images with no lexicon.

\noindent\textbf{ICDAR 2017 MLT(MLT-2017)}~\cite{nayef2017}
is comprised of 9000 training images and 9000 test images. We acquire cropped word instances for recognition by using the quadrilateral word-level annotation.

\noindent\textbf{SynthText}~\cite{synthtext} consists of 80k images for training. We cropped about 7 million instances with character and word-level bounding-boxes annotations from the training set.

\noindent\textbf{Synth90k}~\cite{mjsynth} contains 8 millions word images from 90k English words with word-level annotation.

In addition, approximately 16k real images are collected from the training sets released by the mentioned datasets for fine-tuning.

\subsection{Training Strategy}
The training procedure of TextScanner includes two stages: we firstly use the synthetic dataset SynthText with character-level annotations to pre-train the model for $5$ epochs, then the real image examples with sequence-level annotations are mixed into the training set for fine-tuning the network for $1$ more epoch. Our methods in Tab.\ref{tab:performance} use different data for training, ``TextScanner-pre'' and ``TextScanner-mutual'' are the models trained on SynthText with and without mutual-supervision, respectively; ``TextScanner+90k'' and ``TextScanner+real'' are fine-tuned with the 90k dataset and the training set of real data.

We use Adam optimizer for training of all experiments. The learning rate is initialized as $10^{-3}$ and the decays to $10^{-4}$ and $10^{-5}$. During training and inference, the input images are resized to $64\times256$.

\subsection{Recognition Performance Evaluation}

\begin{figure}[!ht]
\centering
\includegraphics[width=0.97\linewidth]{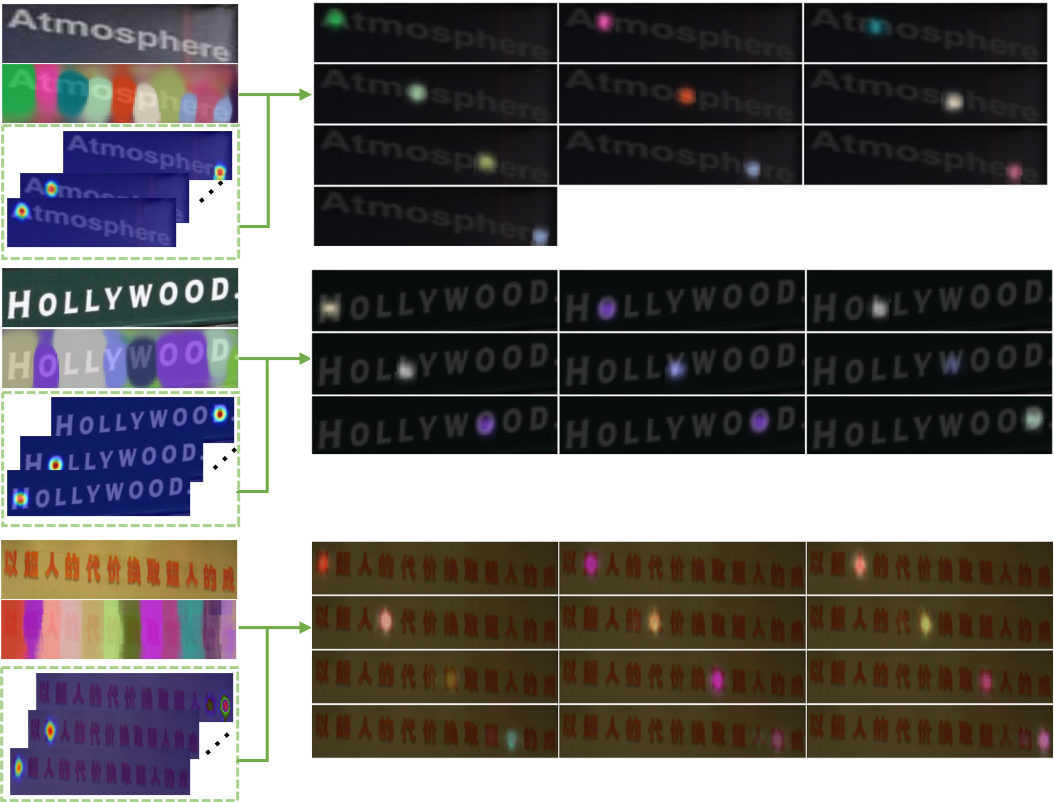}
\caption{Intermediate results of TextScanner. Obviously TextScanner can track the arrangement of characters well, for long or oriented text. The geometry branch can separate adjacent characters even their segmentation masks connect to each other (note the two adjacent `O' the middle row).}
\label{fig:example}
\end{figure}

The recognition accuracies of different methods on standard benchmarks, including regular (IIIT, SVT, IC13) and irregular  (IC15, SVTP, CT) text, are shown in Tab.~\ref{tab:performance}. 

The natural modeling of TextScanner makes it more robust to hard cases where the text is curved or oriented. The three variants of TextScanner in Tab.~\ref{tab:performance} consistently outperform previous methods in comparison with the same training data.
Especially on curved text, ``TextScanner+90k'', which is trained with synthetic data, achieves an improvement of $3.3\%$ on IC15, $4.1\%$ on SVTP, and $4.0\%$ on CT.
The advantages of TextScanner stem from aspects: (1) It is segmentation based, which makes the prediction more relevant to visual features and free from error accumulation brought by the recursive modeling. (2) It scans characters one by one and ensures they are read in the right order and separated properly. Some recognition examples are visualized in Fig.~\ref{fig:example}.

``TextScanner+real'' brings an even more significant boost in performance and demonstrates the capability of TextScanner to utilize real data for training, which also verifies the effectiveness of the proposed mutual-supervision mechanism. Moreover, although the mutual-supervision mechanism is designed for making use of real-world data with sequence-level annotations, fine-tuning with synthetic SynthText and 90k datasets consistently bring performance improvement.

\subsection{Chinese Recognition Evaluation}

Preceding experiments are based on datasets with English text, whose alphabet is relatively small. To further validate the capability of TextScanner, we conduct experiments on a Chinese recognition task, which is more challenging due to its larger alphabet and more complex visual variations of characters.
We compare the performance of TextScanner in Chinese recognition with two representative methods, CRNN\cite{shi2017end} and ASTER\cite{aster}. We use their open-source implementations for comparison. The models are trained with the same training data, which is generated by the synthetic engine released with SynthText and evaluated on cropped text images from the validation set of MLT-17.

The quantitative results are shown in Tab.\ref{tab:Chinese}. In addition to accuracy, we also evaluate the Normalized Edit Distance of the methods following the ICDAR-2017 competition RCTW-17\cite{shi2017icdar2017}:
\begin{equation}
    Norm = 1 - \frac{1}{N_{t}}\sum_{i}^{N_{t}}{\frac{ED(s_{i}, \hat{s_{i}})}{\max(|s_{i}|, |\hat{s}_{i}|)}}
\end{equation}
where $N_t$ is the number of text instances.

\begin{table}[!t]
\centering
    \begin{tabular}{|c|c|c|}
    \hline
    \textbf{Methods} & Acc(\%) & NED \\
    \hline
    CRNN\tiny\cite{shi2017end} & 59.2 & 0.68 \\
    ASTER\tiny\cite{aster} & 57.4 & 0.69 \\
    TextScanner & 64.1 & 0.75 \\
    \hline
    \end{tabular}
\caption{Results comparison on MLT-17. "NED" is short for normalized edit distance.}
\label{tab:Chinese}
\end{table}

As shown in Tab. \ref{tab:Chinese}, TextScanner outperforms attention decoder with a large margin, $6.7\%$ in recognition accuracy and $0.06$ in normalized edit distance. The results demonstrate that TextScanner can handle such challenging recognition tasks better.

The main reason for this improvement is that TextScanner separate geometry branch from the class branch. For the Chinese recognition task, models are more prone to classification errors due to the much more complex structures and appearances of Chinese characters.

In contrast to attention decoders, the dual-branch architecture of TextScanner is more robust to the problem of error accumulating towards difficulty in classification. The class branch and the geometry branch are optimized individually in pre-training, therefore the extraction of character orders is not affected by the probable errors in classification. In the fine-tuning stage, the accurate order extraction can improve the class branch in return through the mutual-supervision mechanism.

\subsection{Character Localization Accuracy of TextScanner}

\begin{figure}[t]
\centering
\includegraphics[width=0.65\linewidth]{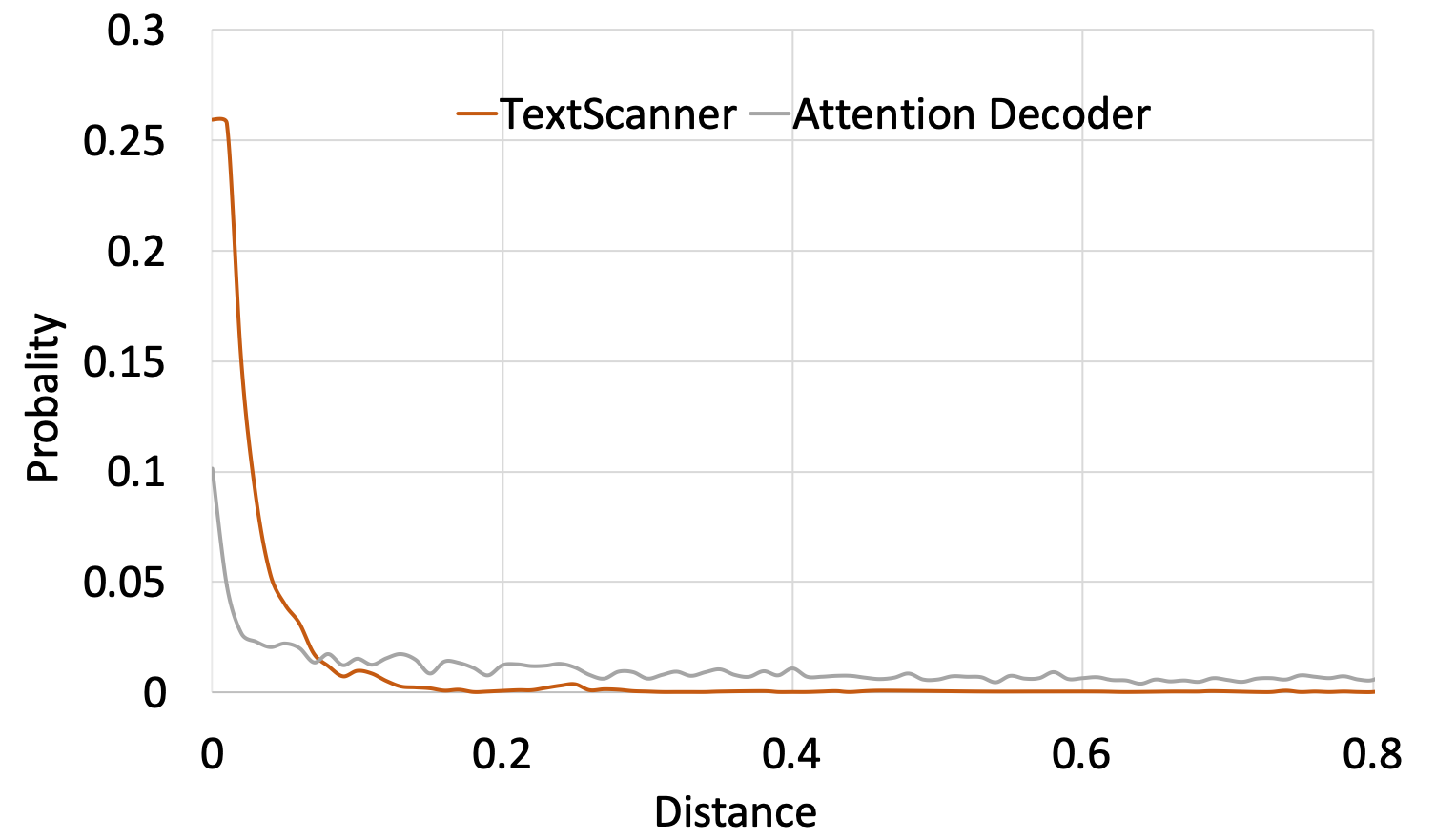}
\caption{Probability density of deviation of character localization. ``Distance'' denotes the distance from predicted character center or attention position to the ground truth character center, which is normalized by the image width.}
\label{fig:localization}
\end{figure}

For both attention decoder and TextScanner, accurate prediction of attention position or character localization is fundamental for recognition. As they both produce the center of characters, we compare their performance in character localization on the IC13 dataset. As IC13 provides annotations of character positions in the image, the two methods are evaluated by measuring the normalized distance $D$ between the produced character center and ground truth center position in the width axis.

The probability density of center distance in IC13 is illustrated in Fig.\ref{fig:localization}. The probability of TextScanner to have accurate localization($D < 0.1$) is obviously greater than attention decoder. This proves that TextScanner gives more accurate character localization results.

\section{Ablation Study}

\begin{table}[ht]
    \centering
    \resizebox{.95\columnwidth}{!}{
    \begin{tabular}{|c|c|c|c|c|c|c|c|}
    \hline
        \multirow{2}{*}{Geo} & \multirow{2}{*}{Ord} & \multicolumn{6}{|c|}{Accuracy ($\%$)}\\ \cline{3-8}
         & & IIIT & SVT & IC13 & IC15 & SVTP & CT \\ \hline
        $\times$ & $\times$ &   90.6    &  84.7   & 90.5 & 72.6 & 73.5 & 81.9 \\ \hline
        $\checkmark$ & $\times$ & 91.2 & 85.6 & 90.9 & 75.8 & 75.6 & 82.7 \\ \hline
        $\checkmark$ & $\checkmark$ & 92.6 & 89.5 & 92.5 & 80.1 & 83.7 & 85.2 \\ \hline
    \end{tabular}
    }
    \caption{Recognition performance with different settings. ``Geo'' denotes the geometry branch, ``Ord'' denotes word formation using the order maps.}
    \label{tab:geometry}
\end{table}

The superiority of TextScanner has been validated via the experiments in Chinese recognition task, here we specifically explore the effect of the geometry branch and the word formation process based on order maps. In Tab.\ref{tab:geometry}, the geometry branch and word formation are ablated respectively. Note that the word formation with order maps relies on the output of the geometry branch, therefore we use the post-processing procedure of CA-FCN as a replacement.

The experimental results clearly show the improvements brought by the geometry branch and its decoding process (the second and the third row in Tab.\ref{tab:geometry}). As the order maps ensure the characters are scanned in correct order, the recognition performance is significantly elevated, especially on irregular datasets: $7.4\%$ on IC15 and $10.2\%$ on SVTP. Besides, even with the regular post-processing, the geometry branch still achieves better performance, proving it can facilitate the optimization of the class branch.

\section{Conclusion}

In this paper, we have presented TextScanner, an effective segmentation-based dual-branch framework for scene text recognition. TextScanner can overcome the problems and defects of previous methods, and work well under various challenging scenarios. A novel mutual-supervision mechanism, which makes it possible to take full advantage of both real and synthetic data, is also proposed. Besides, TextScanner shows stronger adaptability in handling difficult text.
\subsection*{Acknowledgements}
This work was partly supported by National Natural Science
Foundation of China (61733007).

{\fontsize{9.0pt}{10.0pt} \selectfont
\bibliographystyle{aaai}
\bibliography{reference}
}
\end{document}